\documentclass[runningheads]{llncs}

\usepackage{graphicx}
\usepackage{amsmath,amssymb}
\usepackage{booktabs}
\usepackage{multirow}
\usepackage{tabularx}
\usepackage{array}
\usepackage{ragged2e}
\usepackage{xcolor}
\usepackage{comment}
\usepackage[hidelinks]{hyperref}

\graphicspath{{./figures/}}

\begin{document}

\title{RoboSurg-VQA: A Multimodal Benchmark for Surgical Segmentation-Aware Visual Question Answering}
\titlerunning{RoboSurg-VQA}

\author{
Chengyi Zhang\inst{1} \and
Zi Ye\inst{2} \and
Ziyang Wang\inst{3}}

\authorrunning{C. Zhang et al.}

\institute{Swansea University, UK \\ \and
Maynooth University, Ireland \and 
Aston University, UK
}

\maketitle

\begin{abstract}
Reliable visual understanding in robot-assisted and minimally invasive surgery (RMIS/MIS) demands more than accurate masks: in clinical practice, clinicians pose language-like questions about procedural context, visibility, artefacts, and the presence of anatomical structures and surgical instruments, often under degraded views caused by occlusion, smoke, bleeding, and specular highlights. We present \textbf{RoboSurg-VQA}, a segmentation-aware visual question answering (VQA) benchmark built by repurposing public surgical segmentation datasets under a shared schema. Each frame is paired with a fixed set of clinically motivated questions spanning procedure context, anatomy (including region), imaging modality/view, surgical artefacts, image quality, and basic visibility and spatial attributes, with closed answer sets to enable consistent evaluation.
To scale annotation, we generate candidate answers via constrained prompting with automatic validity and consistency checks, followed by human auditing to improve plausibility and label consistency. We report benchmark statistics, sanity baselines, and common evaluation challenges under challenging surgical conditions. The code will be available on \textcolor{pink}{\url{https://github.com/ziyangwang007/Robosurg-VQA}}.


\end{abstract}
\justifying

\section{Introduction}
\label{sec:intro}
Robot-assisted and minimally invasive surgery relies heavily on endoscopic vision, yet restricted field-of-view, specular highlights, smoke, and frequent occlusions make robust scene understanding difficult \cite{alabi2025mtl,wang2025s4roboformer}. The EndoVis 2017 instrument segmentation challenge \cite{allan2019endovis17} and EndoVis 2018 scene segmentation challenge \cite{allan2020endovis18} have established strong foundations for pixel-level surgical perception. Large-scale surgical video datasets annotated with workflow labels have also supported progress in surgical scene understanding \cite{twinanda2017endonet}. However, segmentation leaderboards do not capture the queries clinicians actually ask at the console about context, visibility, and artefacts under uncertainty.

Visual Question Answering (VQA) offers a natural interface for linking perception and language \cite{antol2015vqa,goyal2017vqav2}, but surgical VQA differs from natural-image VQA: terminology is specialised, visual degradation is common, and clinically relevant answers often depend on fine-grained cues. Medical VQA benchmarks such as VQA-RAD \cite{lau2018dataset} and SLAKE \cite{liu2021slake} have further highlighted the need for clinically grounded question-answer pairs. Prior work has introduced structured surgical QA \cite{seenivasan2022surgical}, extended it with localisation \cite{bai2023surgicalvqla}, and more recently explored multimodal instruction and dialogue for surgery \cite{schmidgall2024gpvls,seenivasan2023surgicalgpt,wang2025endochat}. Nevertheless, existing datasets are typically tied to limited sources and label spaces or prioritise dialogue objectives that are not explicitly aligned with \emph{segmentation-aware} evaluation (e.g., visibility/occlusion, artefacts, image quality, and region-sensitive anatomy/instrument cues supported by masks).
\begin{figure}[t]
    \centering
    \includegraphics[width=0.82\linewidth]{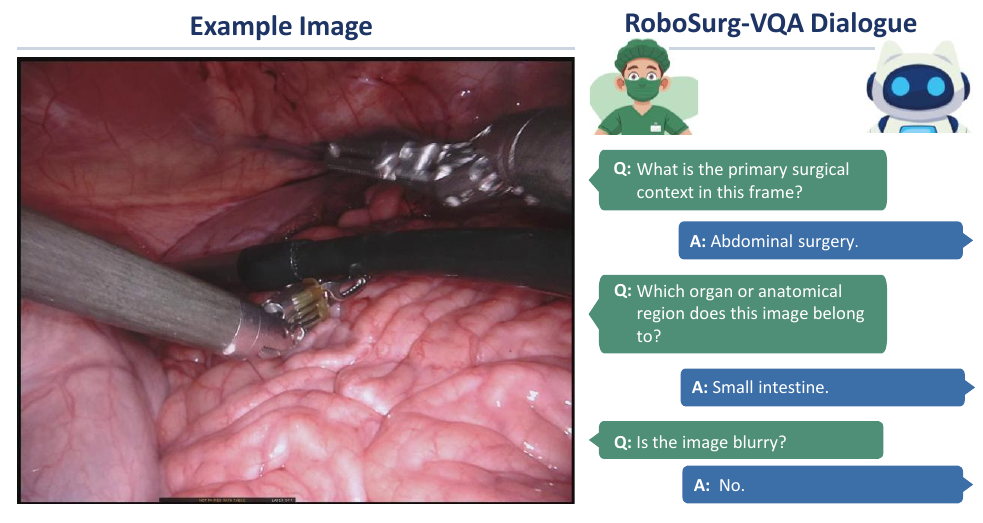}
    \caption{Example dialogue in RoboSurg-VQA (single frame with fixed questions and closed-set answers).}
    \label{fig:dialogue}
\end{figure}
Motivated by recent progress in promptable segmentation \cite{kirillov2023sam,ma2024segment} and text-conditioned medical image segmentation \cite{koleilat2024medclip,zhao2025sat}, we ask whether supervision already available in public surgical segmentation datasets can be converted into a standardised VQA benchmark for evaluating multimodal models under realistic surgical conditions. We introduce \textbf{RoboSurg-VQA}, which unifies heterogeneous sources under a shared schema and pairs each frame with a fixed nine-question taxonomy targeting procedural context, anatomy (including region), imaging modality/view, surgical artefacts (e.g., bleeding, smoke and occlusion), image quality, and basic visibility and spatial attributes. Candidate answers are produced using constrained prompts with automatic validity and consistency checks, and then audited using additional consistency rules to improve reliability.

\begin{figure}[t]
  \centering
  \includegraphics[width=\linewidth]{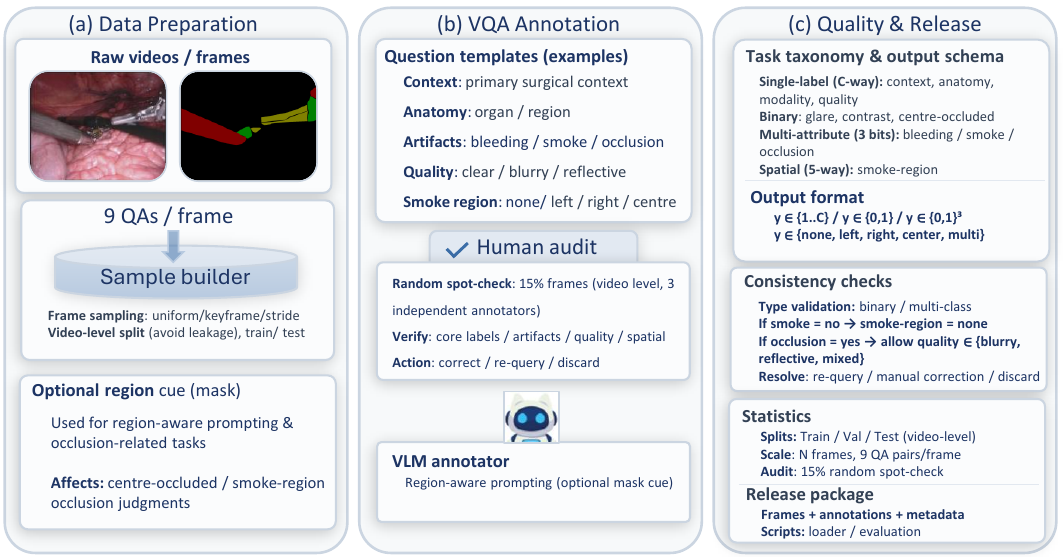}
  \caption{Dataset construction workflow of RoboSurg-VQA, including data preparation, model-assisted VQA annotation with human auditing, and quality control prior to release.}
  \label{fig:workflow}
\end{figure}

In summary, we make four contributions: (i) introducing \textbf{RoboSurg-VQA}, a segmentation-aware VQA benchmark for robotic and endoscopic surgery, built by repurposing public surgical segmentation datasets under a unified schema; (ii) defining a fixed nine-question, closed-set taxonomy spanning context, anatomy (including region), imaging modality, artefacts, image quality, and visibility; (iii) proposing a scalable, constraint-driven annotation and quality-control pipeline with human auditing; and (iv) providing benchmark statistics, sanity baselines, and evaluation protocols to enable reproducible comparison under realistic degradations.

\section{Overview of RoboSurg-VQA}
\label{sec:format}

\subsection{Task Definition \& Benchmark Scope}
RoboSurg-VQA targets \emph{segmentation-aware} VQA for robotic and endoscopic surgery. Given a surgical image \(\mathbf{I}\) and segmentation annotations \(\mathbf{M}\), the task is to answer a fixed set of questions \(\{\mathbf{Q}_k\}_{k=1}^{K}\) using closed answer spaces specific to each question type. This design supports reproducible scoring while grounding questions in surgical evidence and segmentation-relevant cues.

The benchmark covers global scene properties, mid-level attributes, and region-sensitive attributes such as visibility, occlusion, and artefact location. For region-dependent questions, masks provide optional spatial support for reasoning over specific image regions.

\subsection{Source Datasets \& Data Unification}
\begin{table}[!t]
\centering
\caption{Public surgical segmentation sources used in RoboSurg-VQA (robotic and non-robotic endoscopy/laparoscopy). We unify heterogeneous modalities and label spaces into a shared frame-based format for VQA conversion.}
\label{tab:dataset_summary}
\scriptsize
\setlength{\tabcolsep}{3.2pt}
\renewcommand{\arraystretch}{1.05}
\begin{tabularx}{\linewidth}{@{}
>{\raggedright\arraybackslash}p{0.20\linewidth}
>{\raggedright\arraybackslash}p{0.25\linewidth}
>{\raggedright\arraybackslash}p{0.13\linewidth}
>{\raggedright\arraybackslash}X
>{\raggedleft\arraybackslash}p{0.10\linewidth}
@{}}
\toprule
\textbf{Dataset} & \textbf{Domain / Surgery} & \textbf{Modality} & \textbf{Seg. labels} & \textbf{Scale} \\
\midrule
EndoVis17 \cite{allan2019endovis17} & Robotic (porcine), nephrectomy & Stereo RGB & Tool (MC, 7) & $\sim$3.0k \\
EndoVis18 \cite{allan2020endovis18} & Robotic (porcine), training & Stereo RGB & Tool parts + objects (MC) & $\sim$5.7k \\
EndoVis15 \cite{bodenstedt2018comparative} & Robotic, ex-vivo & 2D RGB & Tool (seg.) & $\sim$9.1k \\
SAR-RARP50 \cite{psychogyios2023sar} & Robotic, RARP & Stereo RGB & Semantic masks (MC) & $\sim$16.3k \\
Kvasir-Instrument \cite{jha2021kvasir} & GI endoscopy & 2D RGB & Tool (binary) & $\sim$0.6k \\
Endoscapes23-Seg50 \cite{murali2023endoscapes} & Laparoscopy, cholecystectomy & 2D RGB & Anatomy+tool (MC, 6) & $\sim$0.5k \\
CholecInstanceSeg \cite{alabi2025cholecinstanceseg} & Laparoscopy (human), cholecystectomy & 2D RGB & Tool inst. (MC, 7) & $\sim$41.9k \\
DSAD \cite{carstens2023dresden} & Robotic (human), rectal surgery & 2D RGB & Anatomy (MC, 11) & $\sim$13.2k \\
\bottomrule
\end{tabularx}

\vspace{2pt}
{\scriptsize
\parbox{\linewidth}{\textbf{Notes.}
\textbf{Modality:} 2D RGB and Stereo RGB denote monocular and stereo endoscopic RGB views. \par
\textbf{Seg. labels:} binary denotes tool vs.\ background, MC denotes multi-class segmentation, and parentheses indicate the number of annotated classes/tools when available.}}
\end{table}

RoboSurg-VQA repurposes public surgical image and video datasets with pixel-level annotations for instruments, anatomy, or surgical scenes. We standardise heterogeneous sources into a unified frame-based representation that preserves each frame, its original segmentation mask, auxiliary metadata, and, where available, the original splits to reduce leakage.

This representation maps segmentation-derived information to predefined VQA questions and answer spaces while retaining masks as optional spatial cues for region-dependent questions. Anatomy masks can support anatomy queries when available; in our EndoVis17/18 instantiation, instrument-only masks are used for localisation and visibility cues.

\subsection{VQA Design \& Question Taxonomy}
The VQA component uses fixed question templates with predefined, closed answer spaces to reduce linguistic ambiguity and support consistent benchmarking. The taxonomy covers procedure context, anatomy, imaging setup, artefacts, image quality, visibility, occlusion, and spatial attributes aligned with segmentation cues.

\begin{figure}[!t]
  \centering
  \includegraphics[width=\linewidth]{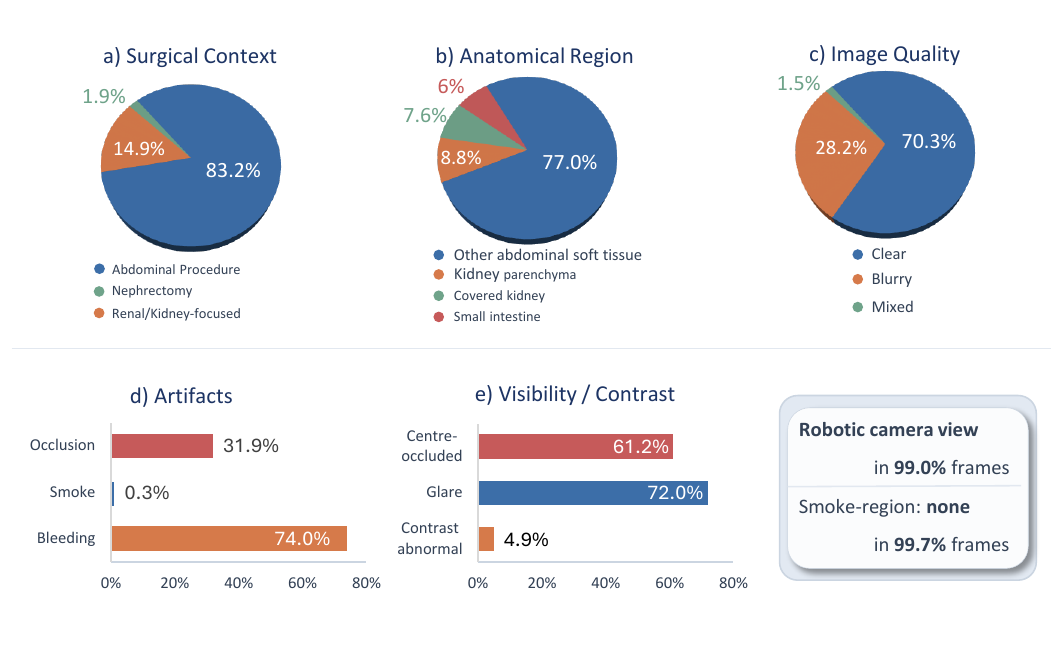}
  \caption{Distribution of question-answer labels in the RoboSurg-VQA benchmark.}
  \label{fig:Distribution}
\end{figure}

Figure~\ref{fig:Distribution} summarises answer distributions, and Table~\ref{tab:robosurgvqa_questions} lists the question templates and answer spaces.

\begin{table}[!t]
\centering
\caption{RoboSurg-VQA question list and controlled answer spaces.}
\label{tab:robosurgvqa_questions}
\scriptsize
\setlength{\tabcolsep}{3.2pt}
\renewcommand{\arraystretch}{1.08}
\begin{tabularx}{\linewidth}{@{}
>{\raggedright\arraybackslash}p{0.20\linewidth}
>{\raggedright\arraybackslash}p{0.30\linewidth}
>{\raggedright\arraybackslash}X
@{}}
\toprule
\textbf{Question (ID)} & \textbf{Question text} & \textbf{Answer space} \\
\midrule

Context (Q1) &
What is the primary surgical context in this frame? &
\textbf{Type:} 3-way categorical (\textit{no Unknown}).\newline
\textbf{Labels:} nephrectomy (robot-assisted), renal/kidney-focused procedure, and other abdominal procedure \\

Anatomy (Q2) &
Which organ or anatomical region does this image belong to? &
\textbf{Type:} 4-way categorical (\textit{no Unknown}).\newline
\textbf{Labels:} small intestine, kidney (parenchyma), covered kidney (fat/fascia-covered kidney), and other abdominal soft tissue (peritoneum/mesentery/fat) \\

Modality (Q3) &
What is the imaging modality/view? &
\textbf{Type:} 4-way categorical.\newline
\textbf{Labels:} endoscopic, robotic camera, synthetic, and Unknown \\

Artifacts (Q4) &
Is there bleeding, smoke, or occlusion visible in the image? &
\textbf{Type:} 3-bit multi-attribute.\newline
\textbf{Fields:} bleeding $\in \{$yes, no$\}$, smoke $\in \{$yes, no$\}$, and occlusion $\in \{$yes, no$\}$ \\

Visual quality (Q5) &
How is the visual quality of the current scene? &
\textbf{Type:} 5-way categorical.\newline
\textbf{Labels:} clear, blurry, reflective, mixed, and Unknown \\

Glare (Q6) &
Is there specular reflection/glare? &
\textbf{Type:} 3-way categorical.\newline
\textbf{Labels:} yes, no, and Unknown \\

Contrast (Q7) &
Is the image contrast normal? &
\textbf{Type:} 3-way categorical.\newline
\textbf{Labels:} yes, no, and Unknown \\

Centre-occluded (Q8) &
Is the centre of the frame occluded by instruments? &
\textbf{Type:} 3-way categorical.\newline
\textbf{Labels:} yes, no, and Unknown \\

Smoke region (Q9) &
Which regions contain smoke? &
\textbf{Type:} 5-way spatial categorical (\textit{strict}).\newline
\textbf{Labels:} none, left, right, centre, and multi \\

\bottomrule
\end{tabularx}
\end{table}

\section{Dataset Statistics and Analysis}

\subsection{Experimental Setup and Model-assisted Annotation}
RoboSurg-VQA uses a shared manifest format, where each sample links an RGB frame, its segmentation mask, site, split, and identifier. To avoid ambiguity across partitions, we treat \((\texttt{site}, \texttt{split}, \texttt{id})\) as the global sample key.

\noindent\textbf{Instantiated benchmark.}
We report an end-to-end instantiation on two public surgical segmentation datasets, totalling 11,480 frames. We split the dataset into 8,745 training and 2,735 test frames across three sites (site1: 2,400; site2: 3,232; site3: 5,848). These statistics are computed from the dataset manifest (to be released upon acceptance) and are used for all subsequent analyses in this paper.

\noindent\textbf{Visual inputs and mask semantics.}
Each query includes two vision inputs: (i) the RGB frame and (ii) a pseudocolour \emph{instrument-only} segmentation mask with no organ classes. The mask provides a spatial cue for instrument localisation and visibility-related attributes (e.g., centre occlusion), rather than semantic supervision for anatomy or procedure context.

\begin{figure}[t]
  \centering
  \includegraphics[width=\linewidth]{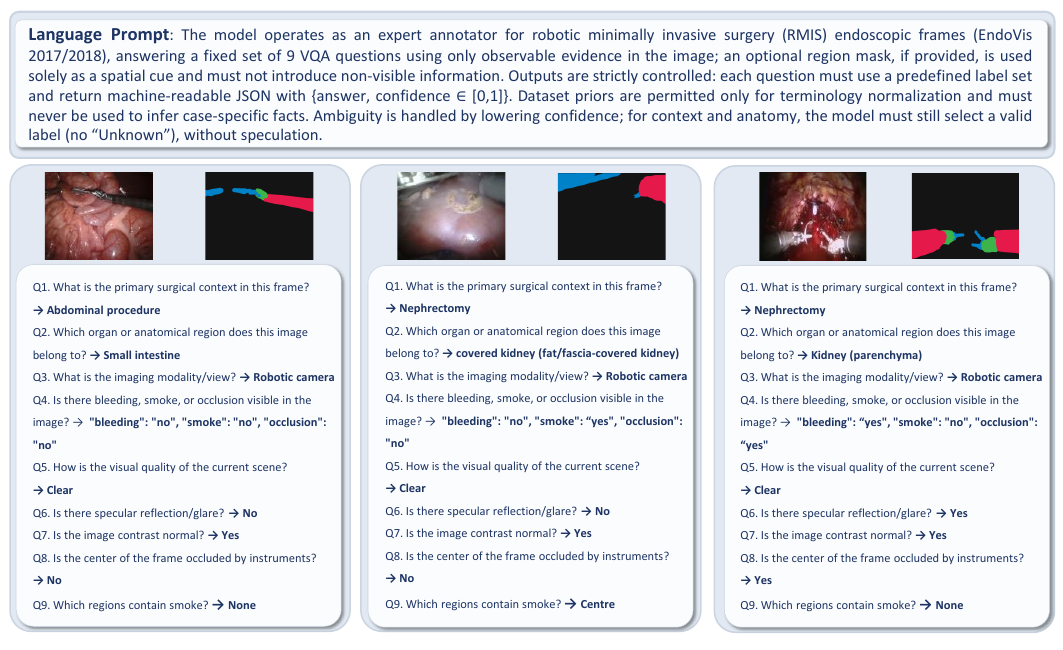}
  \caption{Example of model-assisted VQA annotation in RoboSurg-VQA.}
  \label{fig:Example}
\end{figure}

\noindent\textbf{Constrained annotation protocol.}
We employ model-assisted annotation with decoding (\texttt{temperature}\,$=\,0$) and JSON-only responses.
The vision-language model (VLM) annotator answers a fixed nine-question schema (Q1--Q9) using controlled vocabularies.
For each question, it outputs \texttt{value} and \texttt{confidence} \((\in[0,1])\).
Outputs for Q1 (3-way) and Q2 (4-way) are post-processed via deterministic normalisation to enforce canonical tokens, including alias mapping, safe fallbacks, and confidence clamping. Examples of the constrained model output are shown in Fig.~\ref{fig:Example}.

\noindent\textbf{Human audit.}
After automatic validation, we audit a random subset of frames sampled at the video level to reduce temporal correlation (Fig.~\ref{fig:workflow}). The audit is conducted by three independent annotators, who verify plausibility and cross-question consistency and then take one of three actions: (i) label correction, (ii) re-querying the model under the same constraints, or (iii) discarding the frame when the view is non-interpretable or the label cannot be resolved reliably. We report the audit rate and the proportions of corrected/re-queried/discarded samples in Sec.~\ref{sec:results_stats}.

\noindent\textbf{Consistency checks.}
We apply automatic validity checks to guarantee machine-readable outputs and enforce a cross-question rule: if Q4.\texttt{smoke} \(=\) ``no'', then Q9 is set to ``none''. We report invalid-output rates and constraint-violation patterns to quantify robustness.

\noindent\textbf{Metrics.}
We report Accuracy and Macro-F1 for all classification questions, using Macro-F1 as the primary metric due to class imbalance.
For Q4, we report (i) bit-wise Macro-F1 averaged over the three attributes and (ii) tuple exact-match (primary).
For Q6--Q8, we report Coverage, defined as \(1-N_{\text{Unknown}}/N\), and Conditional Accuracy on the subset excluding Unknown predictions. Unless otherwise stated, the overall score is the unweighted mean of per-question Macro-F1.
\begin{table}[t]
\centering
\caption{Evaluation protocol (primary in bold).}
\label{tab:metrics_protocol}
\scriptsize
\setlength{\tabcolsep}{3.2pt}
\renewcommand{\arraystretch}{1.05}
\begin{tabularx}{\linewidth}{@{}
>{\raggedright\arraybackslash}p{0.28\linewidth}
>{\raggedright\arraybackslash}p{0.14\linewidth}
>{\raggedright\arraybackslash}X
@{}}
\toprule
\textbf{Group} & \textbf{Qs} & \textbf{Metrics (primary in bold)} \\
\midrule
Closed-set cls.      & Q1,2,3,5,9 & Accuracy; \textbf{Macro-F1}; Balanced Acc. \\
Multi-attr.\ (3-bit) & Q4         & Bit-wise Macro-F1; \textbf{tuple exact-match}. \\
With Unknown         & Q6--8      & 3-way Accuracy; \textbf{Macro-F1}; Coverage; Conditional Acc. \\
\midrule
Overall              & all        & Unweighted mean of per-question \textbf{Macro-F1}. \\
\bottomrule
\end{tabularx}
\end{table}

\subsection{Results and Discussion}
\label{sec:results_stats}

\noindent\textbf{Quality-control statistics.}
On the final corpus, our constrained protocol yields fully machine-readable outputs: 0 records contain parsing errors, 0 are missing answers, and 0 predictions fall outside the controlled vocabularies for the closed-set questions (Q1/Q2/Q4/Q5/Q6--Q9). We further enforce a cross-question consistency rule for smoke and region (Q4.\texttt{smoke}=no \(\Rightarrow\) Q9=none). After applying the automatic checks and any subsequent remediation steps, the released annotations contain 0 such violations.

Beyond syntactic validity, we assess semantic reliability via human auditing. We audit 15\% of frames using stratified video-level sampling to ensure coverage across sites/splits and diverse visual conditions. On the audited subset, 4.8\% were corrected by  human annotators, 8.7\% were re-queried under the same constrained protocol (triggered by automatic flags such as low confidence or ambiguity), and 2.3\% were discarded due to insufficient visual evidence or unusable frames. These outcomes are reported as mutually exclusive at the \emph{final} decision stage per audited frame. All re-queried samples are re-validated by the same automatic checks, and any discarded samples are removed from the released set.

\noindent\textbf{Label distributions.}
The instantiated benchmark exhibits pronounced label skew across several questions. Q3 is dominated by ``robotic camera'' (98.82\%). Q4.\texttt{smoke} is overwhelmingly ``no'' (99.72\%), while Q4.\texttt{bleeding} is frequently ``yes'' (73.99\%). For Q1, the dominant class is ``other abdominal procedure'' (83.22\%), followed by ``nephrectomy (robot-assisted)'' (14.86\%). The remaining samples (1.92\%) belong to ``renal/kidney-focused procedure''. For Q2, ``other abdominal soft tissue'' accounts for 77.03\%, with the remaining mass distributed across kidney (8.84\%), covered kidney (7.57\%), and small intestine (6.56\%). Visibility-related labels are also skewed. Q8 centre-occluded is ``yes'' in 61.18\% of frames, and Q4.\texttt{occlusion} is ``yes'' in 31.86\% of frames. These skews motivate reporting \textbf{Macro-F1} as the primary metric, rather than accuracy alone.

As a sanity check on annotation robustness across foundation models, Table~\ref{tab:model_agreement} reports cross-model agreement against the GPT-4o-mini reference, along with schema validity and the ``Unknown'' rate. These scores quantify consistency with the reference rather than accuracy to human ground truth.

\begin{table}[!htbp]
\centering
\caption{Cross-model agreement (ref: GPT-4o-mini).}
\label{tab:model_agreement}
\scriptsize
\setlength{\tabcolsep}{3.2pt}
\renewcommand{\arraystretch}{1.05}
\begin{tabularx}{\linewidth}{@{}
>{\raggedright\arraybackslash}X
>{\centering\arraybackslash}p{0.23\linewidth}
>{\centering\arraybackslash}p{0.12\linewidth}
>{\centering\arraybackslash}p{0.15\linewidth}
@{}}
\toprule
\textbf{Model} & \textbf{Avg. Macro-F1 (Q1--8)} & \textbf{Valid\%} & \textbf{Unk\% (Q3--8)} \\
\midrule
Gemini 2.5 Flash-Lite & 0.599 & 100.0 & 0.0 \\
Qwen-VL-Plus          & 0.494 & 100.0 & 0.0 \\
Claude Haiku 4.5      & 0.583 & 100.0 & 0.0 \\
GPT-4.1-nano          & 0.569 & 100.0 & 5.8 \\
\bottomrule
\end{tabularx}

\vspace{2pt}
{\scriptsize
\parbox{\linewidth}{\textbf{Notes.}
Macro-F1 measures consistency with the reference (not accuracy to human ground truth).
\textbf{Valid\%} indicates schema-valid parses; \textbf{Unk\%} is the ``Unknown'' rate.}
}
\end{table}

\noindent\textbf{Sanity baselines.}
On the test split, a majority-prior baseline achieves apparently strong accuracies for highly skewed questions (e.g., Q3: 0.9810, Q1: 0.8161, Q2: 0.6680, Q5: 0.7941, and Q9: 1.0000). However, its Macro-F1 remains low when minority classes are systematically missed (e.g., Q1 \(\approx 0.30\) and Q2 \(\approx 0.20\)), confirming that accuracy alone is insufficient. For Unknown-enabled questions (Q6--Q8), a dominant-class predictor achieves chance-level balanced accuracy, motivating balanced metrics alongside coverage reporting.

\noindent Overall, these results indicate that (i) strong priors can inflate naive accuracy, and (ii) visibility- and artefact-related attributes warrant stratified analysis (e.g., conditioning on occlusion/quality) to faithfully characterise performance under degraded surgical views.

\section{Conclusion}
We introduced RoboSurg-VQA for robotic surgery and minimally invasive procedures. It is a segmentation-aware VQA benchmark built from public surgical segmentation datasets. RoboSurg-VQA uses a unified, closed-set schema covering procedure context, anatomy/region, modality/view, surgical artefacts, image quality, and visibility. RoboSurg-VQA is built via a scalable pipeline with constrained model-assisted annotation, automatic checks, and human auditing for reproducible evaluation.

RoboSurg-VQA complements existing segmentation and workflow benchmarks by enabling language-based assessment of segmentation-relevant reasoning under realistic degradations. We release standardised annotations and evaluation protocols for fair comparison, and expect RoboSurg-VQA to accelerate robust surgical vision-language systems.



\bibliographystyle{splncs04}
\bibliography{references}

\end{document}